\documentclass[letterpaper, conference, 10 pt]{ieeeconf}  
\usepackage[left=54pt,right=54pt,top=54pt,bottom=54pt]{geometry}
\IEEEoverridecommandlockouts                              
\usepackage{graphics} 
\usepackage{epsfig} 
\usepackage{mathptmx} 
\usepackage{times} 
\usepackage{amsmath} 
\usepackage{amssymb}  
\usepackage{booktabs} 
\usepackage{cite}
\usepackage{amsfonts}
\usepackage{amsbsy}
\usepackage{placeins}
\usepackage{multirow}
\usepackage{tabularx}

\title{\LARGE \bf
Two Degree of Freedom Adaptive Control for Hysteresis Compensation of Pneumatic Continuum Bending Actuator
}

\author{Junyi Shen$^{1}$, Tetsuro Miyazaki$^{1}$, Shingo Ohno$^{2}$, Maina Sogabe$^{1}$, and Kenji Kawashima$^{1}$
\thanks{* This work has been submitted to the IEEE for possible publication.
Copyright may be transferred without notice, after which this version may
no longer be accessible.}
\thanks{$^{1}$Department of Information Physics and Computing, The University of Tokyo, 7-3-1 Hongo, Bunkyo-Ku, Tokyo, Japan
        {\tt\small tetsuro\_miyazaki@ipc.i.u-tokyo.ac.jp, kenji\_kawashima@ipc.i.u-tokyo.ac.jp}}
\thanks{$^{2}$Innovative Project Planning and Promotion Department, Bridgestone Corporation, 3-1-1 Kyobashi, Chuo-Ku, Tokyo, Japan
        }%
}

\begin{document}

\maketitle
\thispagestyle{empty}
\pagestyle{empty}

\begin{abstract}

Soft robotics, with their inherent flexibility and infinite degrees of freedom (DoF), offer promising advancements in human-machine interfaces. Particularly, pneumatic artificial muscles (PAMs) and pneumatic bending actuators have been fundamental in driving this evolution, capitalizing on their mimetic nature to natural muscle movements. However, with the versatility of these actuators comes the intricate challenge of hysteresis - a nonlinear phenomenon that hampers precise positioning, especially pronounced in pneumatic actuators due to gas compressibility. In this study, we introduce a novel 2-DoF adaptive control for precise bending tracking using a pneumatic continuum actuator. Notably, our control method integrates adaptability into both the feedback and the feedforward element, enhancing trajectory tracking in the presence of profound nonlinear effects. Comparative analysis with existing approaches underscores the superior tracking accuracy of our proposed strategy. This work discusses a new way of simple yet effective control designs for soft actuators with hysteresis properties.

\end{abstract}

\section{INTRODUCTION}

Drawing from its flexible structures with unlimited degrees of freedom (DoF), soft robotics offers a seamless interface between humans and machines \cite{rus2015design}. Advances in this domain, particularly in soft actuators crafted from materials like silicone rubber and shape memory alloys, have catalyzed its growth \cite{AZAMI2019111623, huang2018chasing}. Core to this is the deformation-based diving mechanism, with pneumatic actuation, especially McKibben type pneumatic artificial muscles (PAMs), standing out for mimicking biological muscle behavior and boasting notable force-to-weight attributes \cite{aschemann2013comparison}. Pneumatic bending actuators, characterized by segmented compartments or an elastic continuum, further push the envelope, enabling complex movements \cite{helps2018proprioceptive,zhao2015scalable,laschi2014soft, shen2023trajectory}.

However, while enhancing soft robot competencies, flexible driving mechanisms present control challenges. Achieving precision with soft actuators, especially in motion or pose trackings, becomes elusive \cite{7487695}. The inherent hysteresis, a significant nonlinearity that induces lag in response, impedes accurate and rapid soft robotic response \cite{10.1111/j.1096-3642.1985.tb01178.x}. With their compressibility factor, pneumatic actuators exacerbate this issue \cite{helps2018proprioceptive}.

Addressing hysteresis has led to approaches like model-based control techniques and feedforward hysteresis compensation \cite{zang2017position}. The former treats hysteresis as a dynamic perturbation, employing nonlinear solutions \cite{schreiber2011tracking}, while the latter uses combinations of play operators for reproducing hysteresis properties \cite{xie2018hysteresis}. Both approaches, however, are not foolproof, model-based systems can falter with extreme nonlinearities due to the dependence on linear approximation, and crafting a feedforward compensation model may be computationally taxing. An emergent solution is the model-free high-gain feedback control, which promises to counter nonlinearities \cite{TAN2002753}. However, its adoption in soft robotics remains nascent \cite{8913522}, chiefly due to challenges in designing adaptive algorithms ensuring localized high yet universally low feedback gain levels \cite{5464026}.

Standard practice to combat feedback controller lag incorporates a 2-DoF control system \cite{7074664}. Recent findings indicate that integrating a feedforward differential (D) controller improves control performance in soft actuators with hysteresis \cite{shen2023trajectory}. Nonetheless, the controller's efficiency dwindles with minimal reference change rates, where the hysteresis effects are amplified due to occurrence of dead zones \cite{shen2023trajectory}. The quest for an adaptive feedforward controller that remains consistent efficiency across varied reference motions becomes apparent.

Recent studies spotlight the potential of adaptive high-gain feedback control for pneumatic bending actuators \cite{shen2023trajectory}. Its efficacy surges with an error-driven tuning mechanism that adjusts feedback gains based on reference motion signal features and tracking errors \cite{shen_icra}. A drawback, however, lies in the feedback controller's slower response relative to its feedforward counterpart, this inherent lag, accompanied with the high sensitivity to errors, may cause unsmooth response, fluctuation, and even unstability.

In this study, we utilize a pneumatic continuum bending actuator to execute bend motion tracking with a 2-DoF adaptive control system. This actuator is powered by the PAM mechanism and integrated with an elastic metal plate. To tackle hysteresis, we enhanced the existing method by infusing adaptability into the typically static feedforward component, thus making the adaptive system a 2-DoF structure. The new control method autonomously tuning its gain values according to the motion features in reference signal and error levels. Through experimental comparisons, we demonstrate the elevated tracking efficacy of our proposed 2-DoF system.

Subsequent sections delve into our methods and findings. Section 2 discusses the actuator's design and hysteresis traits. Section 3 outlines the 2-DoF adaptive control system. Section 4 validates our control strategy efficacy through experiments. Section 5 concludes.

\section{Pneumatic Continuum Bending Actuator} 

\subsection{Overall Configuration}

\begin{figure}[tb]
    \centering
\includegraphics[width=\linewidth]{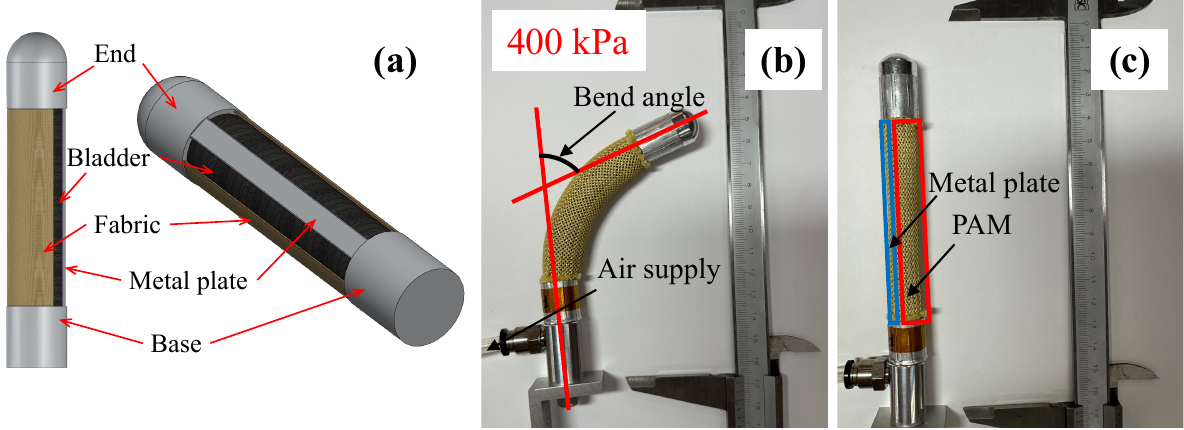}
    \caption{Pneumatic bending actuator: (a) Structural layout; (b) Bending actuation under pressurization; (c) Static straight state at atmospheric level.}
    \label{fig:pam}
\end{figure}

Fig. \ref{fig:pam}(a) illustrates the design schematic of the pneumatic bending actuator. The design encompasses a metallic foundation, a rubber bladder, a metallic tip, a elastic metal plate bridging the base and the tip, and a laterally situated inextensible fabric sheath wrapping the bladder and the plate. The bladder operates on the PAM principle, constrained by the metal plate and the fabric, contracting and exerting a pulling force when pressurized. The incorporation of an elastic metal plate disrupts the uniform radial contraction of PAM, culminating in the actuator's overall bending when subjected to pressurization, as depicted in Fig. \ref{fig:pam}(b). Bending level of the actuator can be modulated by varying its internal pressure values. Upon depressurizing the actuator to atomspheric level, it swiftly reverts to the innate straight configuration, as showcased in Fig. \ref{fig:pam}(b). This response can be attributed to the elasticity of the metal plate.

\subsection{Hysteresis Characteristics}

\subsubsection{Experimental Apparatus}

\begin{figure}[tb]
    \centering
\includegraphics[width=\linewidth]{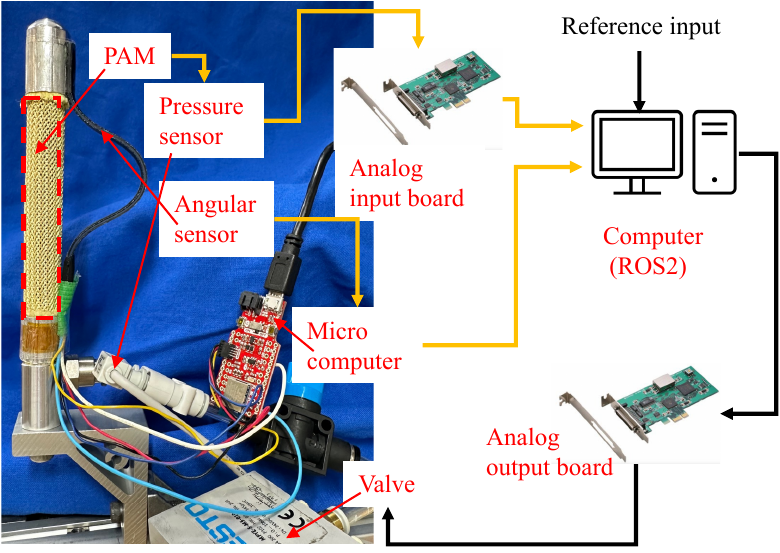}
    \caption{Diagram of the experimental apparatus.}
    \label{fig:exp_app}
\end{figure}

To delineate the hysteresis properties within the soft actuator, we employed the experimental apparatus in Fig. \ref{fig:exp_app}. This configuration amalgamates the bending actuator prototype from Bridgestone, a Festo MPYE-5-M5-010B flow control servo valve, a BENDLABS-1AXIS angular sensor, a SMC PSE 540 A-R06 pressure sensor, and a computer system with a Contec AI-1616L-LPE input board and a Contec AO-1608L-LPE output board, operating within the framework of the robot operating system (ROS2).

\begin{figure}[tb]
    \centering
\includegraphics[width=\linewidth]{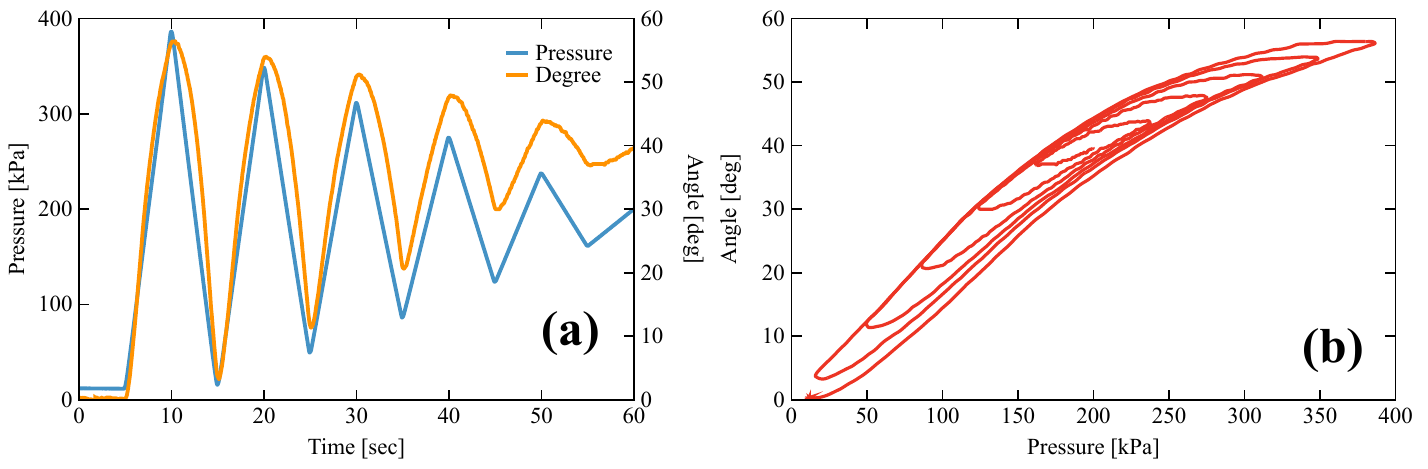}
    \caption{Experimental results: (a) Time-wise variations of pressures and bending angle; (b) Bending angle vs. pressure hysteresis.}
    \label{fig:hys_loops}
\end{figure}

Building on our preceding exploration that mapped pressure-hysteresis relationships using bidirectional pressure waveforms \cite{shen2023trajectory}, this study adopts a streamlined approach \cite{shen_icra}, using a single triangular  waveform that encompasses an extensive pressure spectrum, represented by the blue line in Fig. \ref{fig:hys_loops}(a). Though this sole waveform renders a less granular view of nonlinearity compared to its dual-waveform counterpart, it remains capable in capturing primary hysteresis characteristics of the actuator. Throughout our experiment, records of the internal pressures within the PAM and the bending angles are maintained.

\subsubsection{Analysis of Hysteresis Loops}

The experimental results are exhibited in Fig. \ref{fig:hys_loops}, (a) plots the temporal progression of the PAM's internal pressure and the corresponding bending angle, and (b) unravels the hysteresis between the bending angle and the pressurization level. A feature of this actuator is its asymmetric hysteresis, especially prominent around mid-range bending angles. From hysteresis loops, we observed evident dead zones during the actuator's transition from a pressurized to a depressurized state at the upper side, characterized by a notably broader width compared to the converse transition from depressurization to pressurization at the lower side.

Asymmetric hysteresis property of the bending actuator presents challenges for crafting feedforward hysteresis compensation models. Despite these challenges, the adaptive high-gain control paradigm introduced in \cite{shen2023trajectory}, though originally conceived to cater to a symmetric trend in the widths of hysteresis dead zones against bending levels, remains aptly suitable for this asymmetric scenario, thanks to the versatility in its design.

\section{2-DoF Adaptive High-gain Control System}

\subsection{System Architecture}

\begin{figure*}[tb]
    \centering
\includegraphics[width=\linewidth]{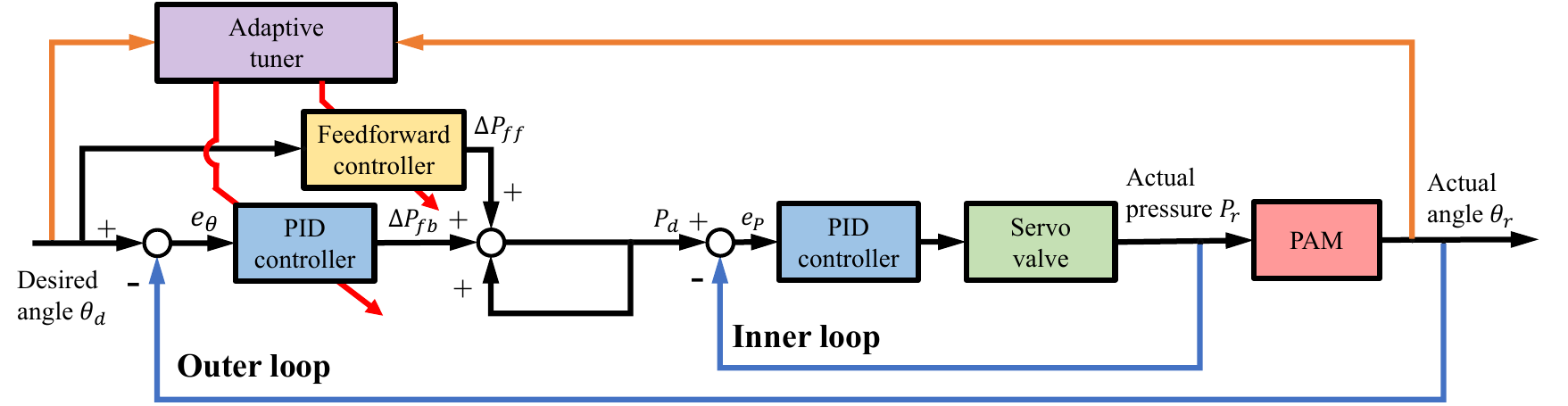}
    \caption{Cascade control system integrating an adaptive anti-hysteresis tuner.}
    \label{fig:system}
\end{figure*}

We design our control system in a discrete manner, using $(t)$ and $(t-i)$ to represent the currently focused, and $i$-th prior discrete instance. The control system, architected in a cascade loop, is illustrated in Fig. \ref{fig:system}. Comprising two loops, the outer loop is for bending angle control and the inner loop is for pressure control, both loops are equipped with a proportional-integral-differential (PID) controller. The feedforward D controller in the outer loop is engineered to counteract the innate delay observed in the soft actuator's response to the reference \cite{chen2019fundamental}. This mitigation is achieved by manipulating the reference angular velocity (termed as the first order pseudo-differential of reference) as:
\begin{equation}
\Delta P_{ff}(t) = K_{ff}(t) \widetilde\theta_d^{(1)}(t),
\label{P_ff}
\end{equation} 
Here, $\widetilde\theta_d^{(i)}(t)$ symbolizes the $i$-th order pseudo-differential of the reference bending angle. For ease of later discussions, the order notation $(i)$ is omitted when equated to $0$. The reference pressure of the inner loop $P_{d}(t)$ is computed by 
\begin{equation}
P_{d}(t) = P_{d}(t-1) + \Delta P_{ff}(t) + \Delta P_{fb}(t), 
\label{P_ref}
\end{equation} 
where $\Delta P_{ff}(t)$ and $\Delta P_{fb}(t)$ represent the pressure adjustment delivered by the feedforward and the feedback controller, respectively.

Critical to the system setup is a 2-DoF adaptive anti-hysteresis tuner that dynamically tailors feedback proportional (P) and feedforward derivative (D) gains in the outer loop. The system's forward paths are shown by black lines, feedback paths are indicated in blue, the orange traces indicate transmission of system states to the tuner, and the red lines exhibit dynamic gain adjustments within the outer loop.

Prior research design the feedforward D component specifically for scenarios where the reference trajectory underwent swift changes with heightened angular velocities \cite{shen2023trajectory}. Conversely, minor reference alterations with low differential values result in a diminished feedforward controller efficacy. To compensate the degenerated feedforward, an adaptive high-gain PID controller was employed \cite{shen2023trajectory}. Though efficient, this adaptive high-gain feedback control is not without its shortcomings. Due to the foundational error-driven mechanism, feedback control, even with escalated gain values, inherently lags in its response relative to its feedforward counterpart. Such lag may result in large tracking errors, which can further result in unsmooth response, fluctuation, or even unstability with the escalated sensitivity to errors feedback gain. Therefore, synergizing a proficient feedforward component with the adaptive feedback controller, particularly during its high-gain phases, could expedite and smoothen the system's response.

\subsection{Anti-hysteresis Adaptive Tuner Design}

The underpinning methodology of our adaptive tuner is inspired by and aligned with the principles laid out in \cite{shen2023trajectory}. Here is a succinct breakdown of its operational paradigm:

For a scenario wherein the desired bending angle displays negative acceleration, denoted as $\widetilde\theta_d^{(2)}(t)$, the gain modulation, symbolized as $\Delta G(t)$, can be articulated by the relationship:
\begin{equation}
\Delta G(t) = M_1 \frac{\widetilde\theta_d(t)|\widetilde\theta_d^{(2)}(t)|}{(b_1+|\widetilde \theta_d^{(1)}(t)|)(c_1+|\widetilde\theta_d^{(2)}(t)|)}D(t).
\label{KP_1}
\end{equation}
In the event where $\theta_d^\text{(2)}(t)$ assumes positive values, the modulation logic morphs to:
\begin{equation}
\Delta G(t) = M_2 \frac{(\Theta-\widetilde\theta_d(t))|\widetilde\theta_d^{(2)}(t)|}{(b_2+|\widetilde\theta_d^{(1)}(t)|)(c_2+|\widetilde\theta_d^{(2)}(t)|)}D(t).
\label{KP_2}
\end{equation}
In any other circumstance, $\Delta G(t)$ is set to zero by default. Within this context, $G(t)$ is indicative of either the proportional gain $K_P(t)$ attributed to the outer loop's feedback PID controller, or the differential gain $K_ff(t)$ associated with the feedforward D controller. The parameters, namely $b$ and $c$, which respectively correlate to velocity and acceleration, undergo meticulous calibration to uphold optimal tracking performance across a spectrum of reference complexities. Furthermore, the directional modulator, $D(t)$, is delineated as:
\begin{equation}
D(t) = -\text{sign}[\widetilde\theta_d^{(1)}(t) \widetilde\theta_d^{(2)})(t)].
\label{D(t)}
\end{equation}

In \cite{shen2023trajectory}, the described adaptation rule is strategically deployed to augment feedback gains during pivotal U-turn motions within the reference trajectory. This enhancement aims to expedite the bending actuator's state transitions, effectively attenuating the resultant hysteresis responses caused by the inherent presence of dead zones. 

In this study, we extend this principle by applying the same adaptive law to the feedforward D gain values. This adaptation is predicated on the hypothesis that a heightened differential gain can sustain the efficacy of the feedforward controller during the absence of high differential values when the reference trajectory decelerates and undergoes directional changes. Such scenarios typically correspond to the actuator's state transitions.

The primary objective behind escalating feedforward gain is to offset the feedback controller's inherent response delays. Doing so allows the system to rapidly and smoothly navigate between depressurization and pressurization phases, mirroring the reference trajectory without considerable lag.

Integrating several constraint functions into the dynamic gain adjustments is imperative to ensure that the gain values are high locally but remain low globally. Existing candidates include the natural-gain-decreasing \cite{shen2023trajectory} function and the error-driven approach \cite{shen_icra}. Although the latter has demonstrated its improved efficacy for feedback gains, we posit that the former might better enhance the performance of the feedforward controller, given its innate open-loop, error-independent nature.

For dynamically adjusting feedback gains, the error-driven function modifies the gain adjustments as follows:
\begin{equation}
\Delta K_P \leftarrow \Delta K_P f(e, D, \kappa) g(K_P, D, \lambda)
\label{stability_fb}
\end{equation}
Here, \(f(e, D, \kappa)\) compresses positive gain adjustments and amplifies negative ones when absolute error is larger than 1:
\begin{equation}
f(e, D, \kappa) = \left(\max\{1, |e|\}\right)^{-\kappa D(t)}
\label{f_theta}
\end{equation}
Additionally, the autonomous gain-central function, \(g(K_P, D, \lambda)\), considers the present magnitude of the proportional gain:
\begin{equation}
g(K_P, D, \lambda) = \left[\frac{K_P(0)}{K_P (t-1)}\right]^{\lambda D(t)},
\label{f_kp}
\end{equation}
which serves to restrain gain growth and quicken its reduction. Within the calculation of \(f(e, D, \kappa)\), the absolute error \(|e|\) is interpreted as a non-dimensional metric. Hence, the exponent \(\kappa\) can be treated as unitless.

On the other hand, for the feedforward D gain, the natural-gain-decreasing function amends its adjustment using:
\begin{equation}
\Delta K_{ff} \leftarrow \Delta K_{ff} h(D, \mu),
\label{stability_ff}
\end{equation}
expressed as:
\begin{equation}
h(D, \mu) = \left(1 + \frac{1 - D(t)}{2} \cdot \mu\right) D(t),
\label{h(t)}
\end{equation}
This effectively magnifies any negative adjustments by \(1+\mu\), ensuring that the gain's reduction pace surpasses its increment rate, thus endowing the dynamic gain with a natural decremental tendency.

To uphold the baseline of the dynamic gains as their initial value \(G(0)\) during updates, we routinely perform a cutoff operation:
\begin{equation}
G(t)=\max\{G(0), G(t-1) + \Delta G(t)\}.
\end{equation}

\section{Experimental Validation}

In this section, we present two sets of comparative experiments to highlight the superior performance of the 2-DoF adaptive high-gain system in mitigating the hysteretic effects during bending trajectory tracking of the pneumatic continuum actuator. For comparison, we benchmark our approach against both the conventional PID controller and the state-of-the-art adaptive high-gain control methods regarding tracking accuracy.

\subsection{Experimental Setup}

Our experimental setup follows the framework illustrated in Fig. \ref{fig:exp_app}. We benchmark the proposed 2-DoF adaptive system (2DoF) against various contemporary methods: the conventional PID, PID augmented with a static feedforward differential element (FF-static), PID complemented by a dynamic feedforward (FF-adaptive), and PID with static feedforward paired with dynamic feedback (FB-adaptive). To rigorously evaluate the performance, we selected two distinct reference bending trajectories: a 30-second trajectory encompassing rapid variations and a longer 120-second trajectory characterized by more gradual changes. These trajectories challenge the controllers to exhibit their proficiency under diverse tracking objectives.

Throughout experiments, we employed a sampling frequency of 500 Hz for data acquisition and signal processing. This frequency remained consistent across all data streams, including the reference and measured bending angles, PAM pressures, and dynamic gain refresh intervals. An important consideration is the adaptability of both the feedback and feedforward gains. Although they can be independently fine-tuned to cater to the asymmetric characteristics of the hysteresis loops, for simplifying the investigation and to reasonably reduce the tuning complexity, we opted for symmetrical parameter settings in this work. As such, parameters such as $M_1$, $M_2$, $b_1$, $b_2$, $c_1$, and $c_2$ from the adaptation program are unified into $M_{ff}$, $M_{fb}$, $b_{ff}$, $b_{fb}$, $c_{ff}$, and $c_{fb}$, respectively. While the symmetric setting might not exploit the full capabilities of both dynamic gains to counteract the asymmetric hysteresis inherent to the pneumatic bending actuator, effectiveness of the adaptive gains and the 2-DoF system is still evident due to their notable resilience against hysteresis even when roughly tuned.

Parameters for the control system were chosen through trial and error. These parameters encompass the gains for the PID controllers across two cascaded loops, the feedforward D gain, and all factors within the 2-DoF adaptive high-gain tuner. These parameters are consolidated and presented in Tables \ref{gain_PID} and \ref{param_adapt} for ease of reference. The settings in these tables are applied to track both the short 30s and the extended 120s bending motion trajecotory.

\begin{table}[htb]
\centering
\caption{Tuned PID Gains for the Cascaded Control Framework}
\label{gain_PID}
\begin{tabular}{lll}
\toprule
Gain & Value & Unit\\
\midrule
Inner - P & $8.0\times10^{-2}$ & V/kPa\\ 
Inner - I & $2.0\times10^{-5}$ & V/kPa$\cdot$ s\\ 
Inner - D & 0 & V$\cdot$ s/kPa \\ 
Outer - P & $1.0\times10^{-1}$ & kPa/deg\\ 
Outer - I & $1.0\times10^{-5}$ & kPa/deg $\cdot$ s\\
Outer - D & 0 & kPa $\cdot$ s/deg\\ 
Feedforward - D & $5.0\times10^{-3}$ & kPa $\cdot$ s/deg\\ 
\bottomrule
\end{tabular}
\end{table}

\begin{table}[htb]
\centering
\caption{Tuned Parameters for the 2-DoF Adaptive Tuner}
\label{param_adapt}
\begin{tabular}{lll}
\toprule
Parameter & Value & Unit\\
\midrule
$M_{fb}$ & $1.4$ & kPa/deg $\cdot$ s\\ 
$b_{fb}$ & $20$ & deg/s\\
$c_{fb}$ & $3.5\times10^{5}$ & deg/$\text{s}^2$\\
$M_{ff}$ & $0.15$ & kPa/deg\\ 
$b_{ff}$ & $30$ & deg/s\\
$c_{ff}$ & $2.0\times10^{5}$ & deg/$\text{s}^2$\\
$\Theta$ & $60$ & deg\\ 
$\kappa$ & $1.0$ & -\\ 
$\lambda$ & $0.20$ & -\\ 
$\mu$ & $0.60$ & -\\ 
\bottomrule
\end{tabular}
\end{table}

\subsection{Experimental Results and Analysis}

\begin{figure}[tb]
    \centering
\includegraphics[width=\linewidth]{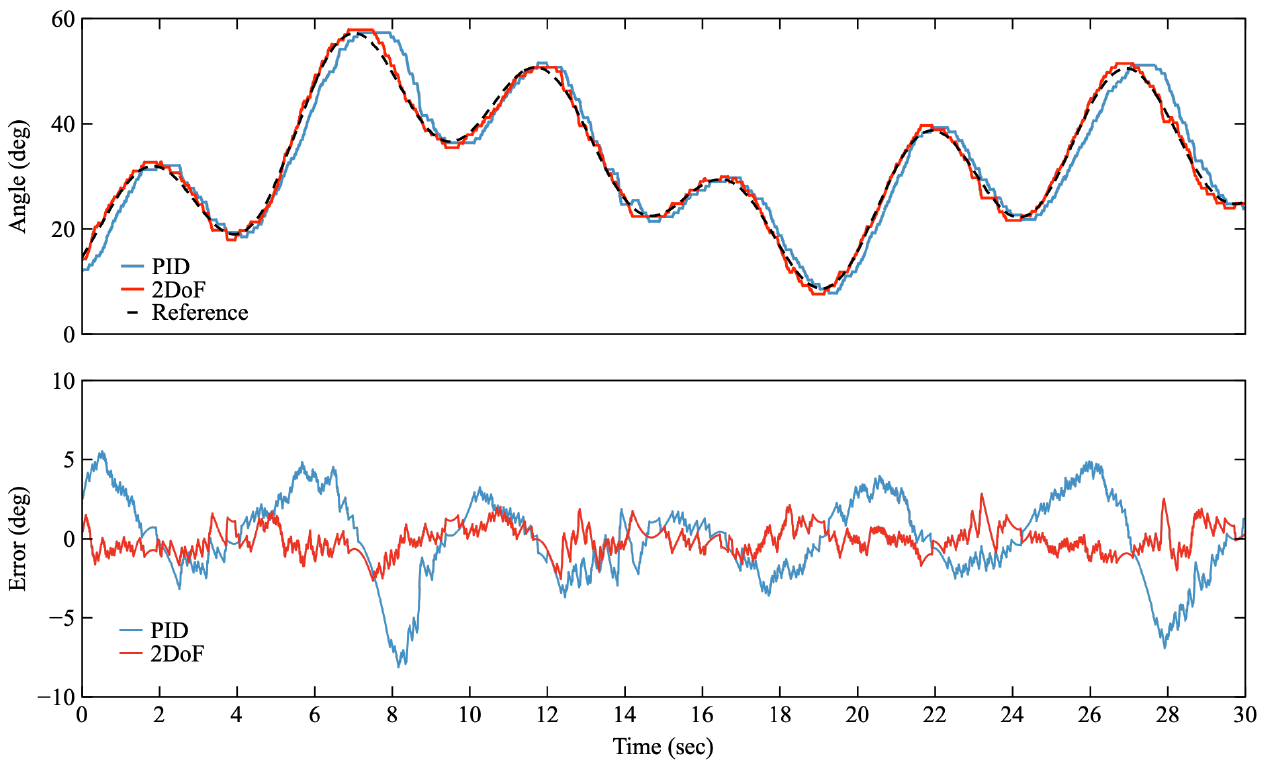}
    \caption{Comparative tracking performance of different control strategies following a 30s reference.}
    \label{fig:short}
\end{figure}

The comparison between the PID and the 2-DoF control methods during the short reference tracking is presented in Fig. \ref{fig:short}. Outcomes stemming from other methods are numerically documented in Table \ref{short_table} and are excluded from the figure for visual clarity. Optimal results within the table are emphasized using bold fonts. The shortfalls of the traditional PID controller in addressing hysteresis become unmistakable. These limitations are particularly pronounced during transitions between pressurization and depressurization phases of the actuator, which align with U-turn motions of the reference trajectory. Integrating a feedforward D element (FF-static )into the control system substantially curtails tracking discrepancies by offsetting system response lags. Nevertheless, it does not notably diminish the peak error magnitudes during U-turn movements. At these junctures, reference differentials descend, resulting in insufficient feedforward element effort, combined with the amplified hysteresis effect in these places, leading to pronounced response lags and consequential tracking inaccuracies. Incorporating adaptability exclusively into either feedback (FB-adaptive) or feedforward  (FF-adaptive) gains leads to commendable enhancements in tracking performance, diminishing more errors than static elements do. Notably, their integration in the 2-DoF adaptive control strategy emerges as the superior technique. The root mean square error (RMSE) and the variation (Var) in error values are reduced by 20\% and 40\%, respectively, compared to standalone adaptive controls. 

\begin{table}[htb]
\centering
\caption{Performance metrics of tracking the 30s bending reference.}
\label{short_table}
\begin{tabularx}{0.48\textwidth}{l*{5}{X}}
\toprule
\multirow{2}{*}{Method} & $e_\text{max}$ [deg] & $e_\text{min}$ [deg] & $|e|_\text{ave}$ [\%] & RMSE [deg] & Var [$\text{deg}^2$] \\
\cmidrule(lr){2-6}
PID & 5.55 & -8.12 & 7.15 & 2.59 & 6.69\\ 
FF-static & 3.49 & -5.19 & 3.74 & 1.46 & 2.12\\ 
FF-adaptive & 3.11 & -3.49 & 3.17 & 1.12 & 1.23\\ 
FB-adapitve & 2.94 & -4.11 & 2.87 & 1.09 & 1.18\\ 
2DoF & \textbf{2.83} & \textbf{-2.70} & \textbf{2.56} & \textbf{0.846} & \textbf{0.706}\\ 
\bottomrule
\end{tabularx}
\end{table}

\begin{figure}[tb]
    \centering
\includegraphics[width=\linewidth]{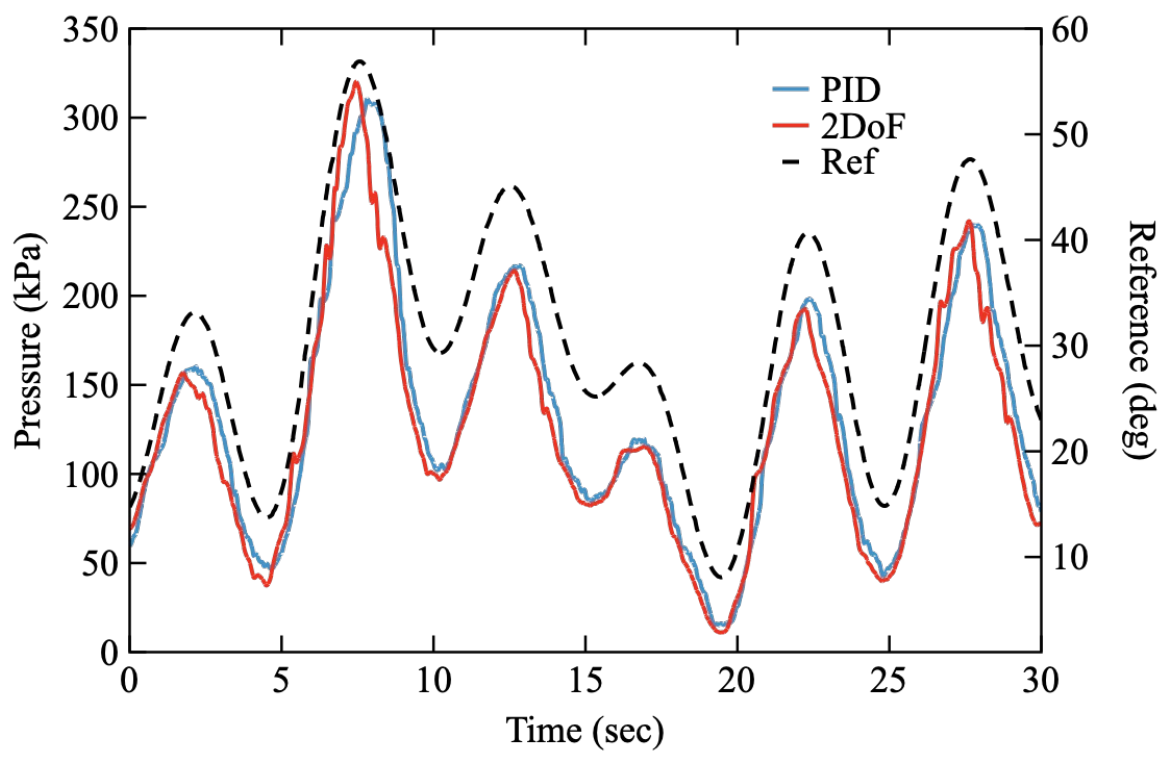}
    \caption{Pressure variations of tracking the 30s bending reference using PID and 2-DoF control method.}
    \label{fig:pressure}
\end{figure}

Fig. \ref{fig:pressure} shows the variation of internal pressure values during motion tracking. The observed reduction in error values at the pneumatic actuator's state transition is attributed to the earlier pressure response in the 2-DoF system, especially during the phase transition from pressurization to depressurization, which exhibits more distinct pressure changes across various control methods than the reverse depressurization-pressurization transition due to the asymmetric hysteresis properties of the actuator, as shown in Fig. \ref{fig:hys_loops} (b). The 2-DoF control introduces fluctuations in internal pressure during state transitions due to elevated gain values. However, these fluctuations appear to minimally impact tracking performance shown in Fig.\ref{fig:short}, likely attributed to the system's insensitivity towards pressure variations within the hysteretic dead zones.

\begin{figure}[tb]
    \centering
\includegraphics[width=\linewidth]{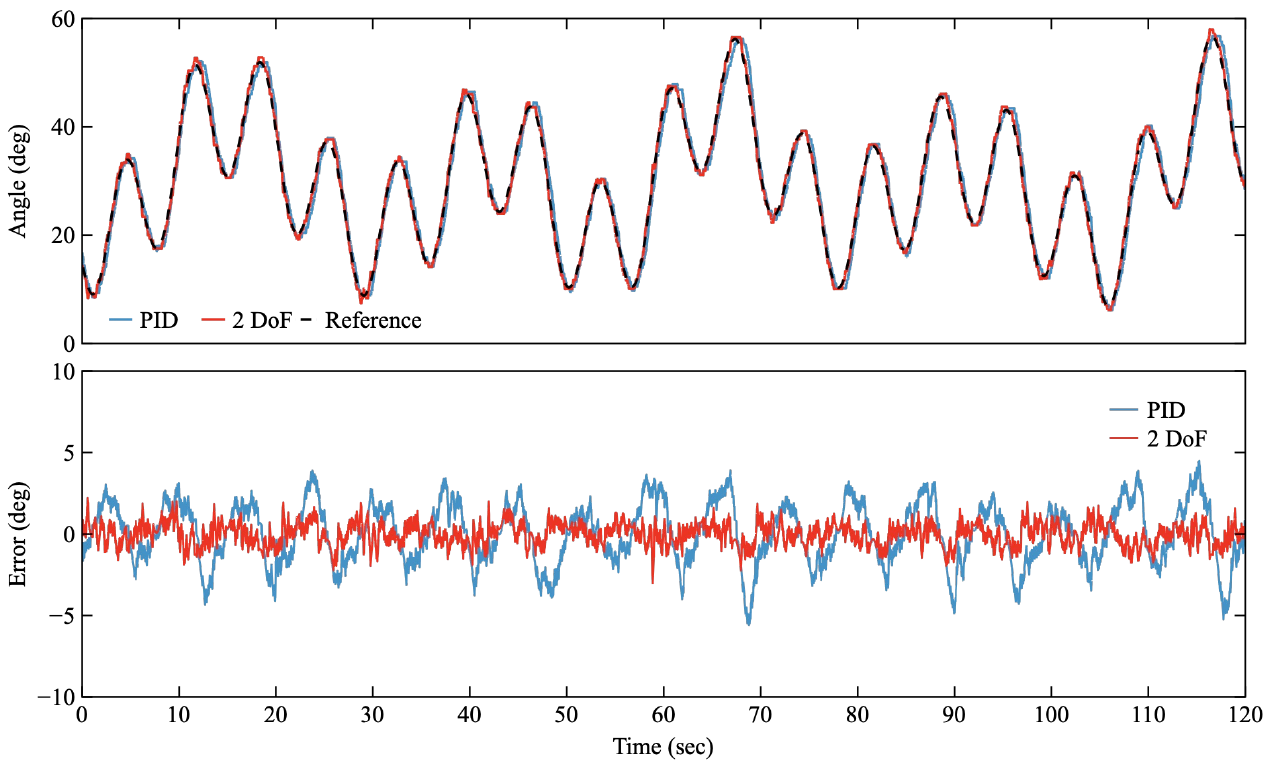}
    \caption{Comparative tracking performance of different control strategies following a 120s reference.}
    \label{fig:long}
\end{figure}

Fig. \ref{fig:long} and Table \ref{long_table} detail the comparative tracking results for the 120s bending motion reference. Various control methods demonstrated better performance for this prolonged reference since it changes less frequently than the former short one. Still, the PID controller struggles with hysteresis, the error pattern is similar to the former short reference. While including a static feedforward mechanism significantly diminishes errors, it marginally affects peak error values. Both the adaptive feedback and adaptive feedforward methods substantially mitigate hysteretic effects, as evident in their reduction of both overall and peak error magnitudes. The 2-DoF adaptive control amalgamates these benefits and emerges superior across all evaluation metrics. RMSE and variation of errors of the 2-DoF method are reduced 15\% and 30\% respctively compared to 1-DoF adaptive approach. Considering the alike error pattern, contraction of the pressure variations across different control method is omitted here for conciseness.

In both reference signal trackings, the 2-DoF adaptive system exhibits no noticeable hysteresis or lag features in their error, as evidenced in Fig. \ref{fig:long} and Fig. \ref{fig:short}. This observation is further corroborated by the minimal error dispersions presented in their respective tables, which indicating exceptional counter-hysteresis capability of the 2-DoF adaptive control method.

\begin{table}[htb]
\centering
\caption{Performance metrics of tracking the 120s bending reference.}
\label{long_table}
\begin{tabularx}{0.48\textwidth}{l*{5}{X}}
\toprule
\multirow{2}{*}{Method} & $e_\text{max}$ [deg] & $e_\text{min}$ [deg] & $|e|_\text{ave}$ [\%] & RMSE [deg] & Var [$\text{deg}^2$] \\
\cmidrule(lr){2-6}
PID & 4.47 & -5.58 & 5.67 & 1.85 & 3.43\\ 
FF-static & 4.33 & -5.36 & 2.83 & 1.02 & 1.03\\ 
FF-adaptive & 2.85 & -3.57 & 2.72 & 0.861 & 0.741\\ 
FB-adapitve & 3.16 & -3.37 & 2.36 & 0.824 & 0.679\\ 
2DoF & \textbf{2.61} & \textbf{-2.43} & \textbf{2.15} & \textbf{0.691} & \textbf{0.476}\\ 
\bottomrule
\end{tabularx}
\end{table}

\begin{figure}[tb]
    \centering
\includegraphics[width=\linewidth]{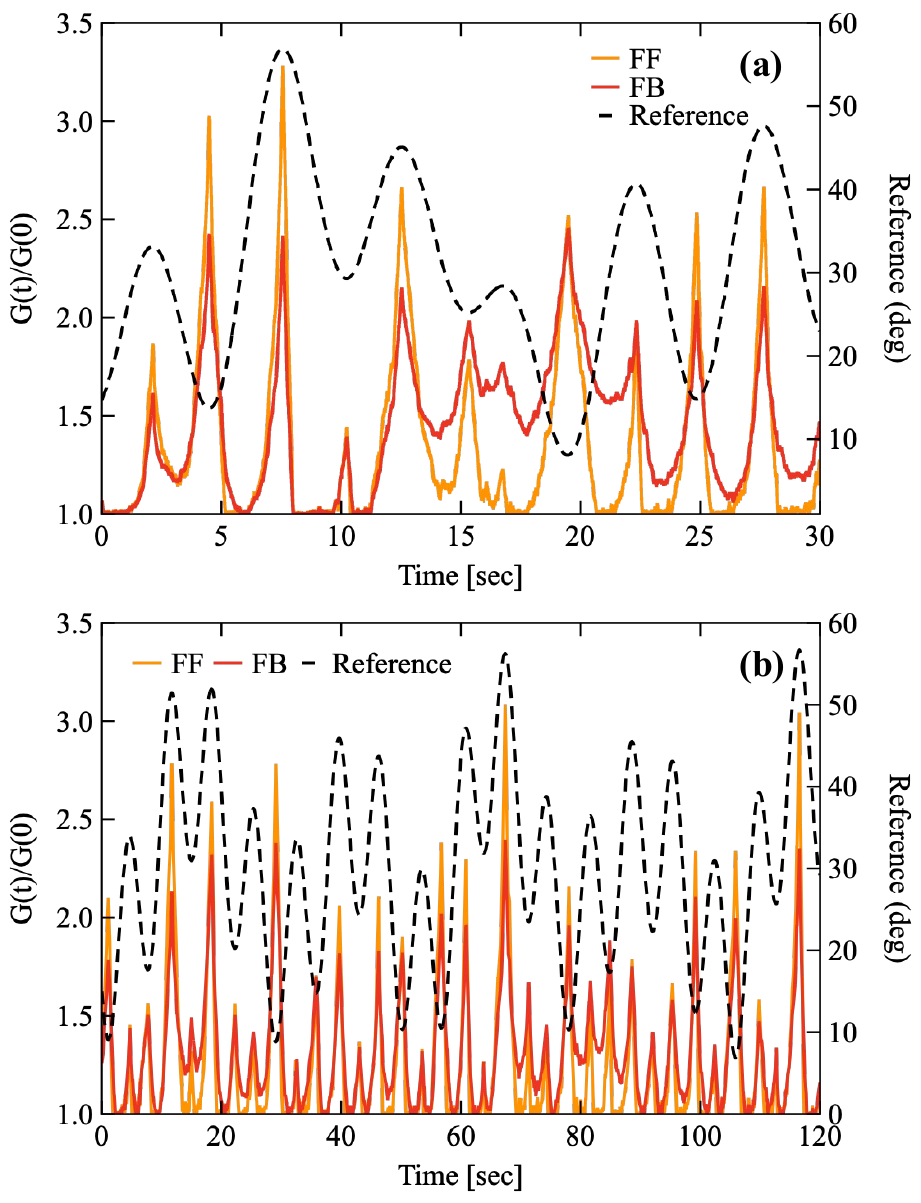}
    \caption{Variation the amplification of dynamic gains in relation to the reference trajectory.}
    \label{fig:ratio}
\end{figure}

In Fig. \ref{fig:ratio}, the dynamic changes in the amplification ratios of feedback proportional (FB) and feedforward derivative (FF) gains within the 2-DoF adaptive system are depicted, with $G(t)$ and $G(0)$ representing the instantaneous and initial dynamic gain values, respectively. Different gains exhibit a similar variation as they increase during U-turn motions due to the shared adaptive rule. Notably, the adaptive feedforward gain peaks higher than the feedback gain as the result of diffrent parameter configurations. The feedback gain's error-driven stabilization mechanism ensures its reduction rate being slower than the feedforward gain that employs a natural-gain-decreasing function. This remained amplification in feedback gain preserves the system's heightened error sensitivity and swift response during low error instances, bolstering bending tracking proficiency. As the system exits pronounced hysteresis zones, the feedforward gain's amplification swiftly regresses to a stable one. The fluctuating amplifications highlight the adaptive rule's versatility; they autonomously augment as the system encounters its hysteresis regions and taper off when the reference undergoes swift changes, ensuring system stability.

\section{Conclusion}

In this work, we achieved precise bending tracking with a pneumatic continuum actuator and a model-free 2-DoF adaptive control system. An experimental analysis elucidated the asymmetric hysteresis within the pneumatic actuator. We employed a cascade loop with a feedforward element to control this actuator. An embedded counter-hysteresis 2-DoF tuner facilitates the autonomous adjustment of feedback and feedforward gain values, contingent on the motion patterns derived from the reference signal. These dynamic gains are modulated based on a shared rule but differ in stabilization functions and parameter specifications due to there different control principle.

Experiments encompassing two distinct bending references validate the efficacy of the 2-DoF adaptive system in mitigating hysteresis during soft actuator manipulation. Comparitive investigation accentuates the superiority of our approach relative to other model-free control approaches. Tracking errors using the 2-DoF adaptive method shows the absence of conspicuous hysteresis features.

This research fortifies advancements in the model-free precision control of hysteresis-riddled soft actuators. The introduced 2-DoF adaptive strategy is distinguished by its uncomplicated nature and remarkable counter-hysteresis proficiency, promising broader applications in soft robotics research faced with hysteresis challenges. 

\addtolength{\textheight}{-12.5cm} 
\bibliographystyle{IEEEtranS}

\end{document}